\documentclass{article}
\usepackage{ijcai22}

\usepackage{times}
\usepackage{soul}
\usepackage{url}
\usepackage[hidelinks]{hyperref}
\usepackage[utf8]{inputenc}
\usepackage[small]{caption}
\usepackage{graphicx}
\usepackage{amsmath}
\usepackage{amsthm}
\usepackage{booktabs}
\usepackage{algorithm}
\usepackage{algorithmic}
\usepackage{subcaption} 
\usepackage{multirow} 
\usepackage{amsfonts} 

\title{Simple and Effective Relation-based Embedding Propagation for \\ Knowledge Representation Learning}

\author{
Huijuan Wang\and
Siming Dai\and
Weiyue Su\and
Hui Zhong\and
Zeyang Fang\and\\
Zhengjie Huang\and
Shikun Feng\and
Zeyu Chen\and
Yu Sun\And
Dianhai Yu\\
\affiliations
Baidu, Inc.\\
\emails
\{wanghuijuan03, daisiming, suweiyue, zhonghui03, fangzeyang, \\
huangzhengjie, fengshikun01, chenzeyu01, sunyu02, yudianhai\}@baidu.com
}

\begin{document}

\maketitle

\begin{abstract}
Relational graph neural networks have garnered particular attention to encode graph context in knowledge graphs (KGs).
Although they achieved competitive performance on small KGs, how to efficiently and effectively utilize graph context for large KGs remains an open problem.
To this end, we propose the \textit{Relation-based Embedding Propagation} (REP) method.
It is a post-processing technique to adapt pre-trained KG embeddings with graph context.
As relations in KGs are directional, we model the incoming head context and the outgoing tail context separately.
Accordingly, we design relational context functions with no external parameters.
Besides, we use averaging to aggregate context information, making REP more computation-efficient. 
We theoretically prove that such designs can avoid information distortion during propagation.
Extensive experiments also demonstrate that REP has significant scalability while improving or maintaining prediction quality.
Notably, it averagely brings about 10\% relative improvement to triplet-based embedding methods on OGBL-WikiKG2 and takes 5\%-83\% time to achieve comparable results as the state-of-the-art GC-OTE.

\end{abstract}

\section{Introduction}\label{sec:introduction}

Knowledge graphs (KGs) are potentially valuable for many applications, such as question answering \cite{KGQA,CQA} and recommender systems \cite{KGRS}.
Although KGs contain rich facts in the form of triplets (\textit{head\ entity, relation, tail\ entity}), they are still far from complete, necessitating a demand for KG completion.
A promising approach is to embed KGs into latent space and predict missing facts from existing ones.
Triplet-based methods \cite{TransE,DistMult,RotatE,OTE} often regard relations as operations over entity embedding space.
They can quickly scale to large KGs because of their elegant simplicity and good interpretability.
However, triplet-based methods overlook global neighbor information.
Recently, considerable literature has grown around graph neural networks (GNNs) to model such graph contexts.
Existing studies for multi-relation KGs \cite{RGCN,CompGCN,TransGCN,M2GNN} usually train relational GNNs from scratch and optimize triplet-based objects.
Specifically, these context-based methods update the central entity with the aggregated embeddings of connected entities and relations.
Thus, the time complexity is proportional to the number of triplets and the average degree of entities.
The high complexity hinders their applications on real-world tasks at a large scale. 
How to efficiently and effectively utilize graph context in large KGs remains a challenge.

To this end, we propose \textit{Relation-based Embedding Propagation (REP)} to combine triplet-based methods' simplicity and graph context strength.
On the one hand, we derive inspiration from simplified GNN methods such as SGC \cite{SGC} and APPNP \cite{APPNP} and remove unnecessary non-linearity.
Instead of parameterizing the aggregation of neighbors with relation-specific transformation, we use triplet-based methods to capture triplet information firstly and then adapt the pre-trained embeddings with graph context.
We do not use layer-specific transform matrices during embedding propagation to reduce time and space complexities.
Our method incorporates graph structures after parameter training and has no activation function or backward propagation during embedding propagation, making it more computationally efficient.

On the other hand, simple embedding propagation that ignores relations, such as SGC and APPNP, is not suitable for KGs and will cause performance degeneration.
Therefore, we propose a novel way of combining neighbor relations and entities.
We make opposite assumptions about the head context and the tail context.
For example, we regard incoming relations as additions and outgoing relations as subtractions.
Then context embeddings are aggregated from neighbor entity-relation pairs.
We take an update scalar to balance pre-trained embeddings and relational context embeddings.
Besides, we provide theoretical and empirical evidence that our proposed method of incorporating relations can help REP obtain an accuracy gain.

We show that REP-OTE takes 5\%-83\% time to achieve comparable performance to the state-of-the-art GC-OTE \cite{OTE} on small datasets through a series of experiments on the link prediction task. 
Our REP also works well on the medium and the large KGs.
In particular, it brings about a 10\% relative improvement on the medium OGBL-WikiKG2 \cite{OGB}, despite taking less than 20\% time for computation.
The key advantages of our REP are three-fold:
\begin{itemize}
    \item By designing the non-parametric embedding propagation without backward propagation, REP becomes computation-efficient and straightforward in utilizing graph context. 
    \item In order to avoid information distortion during embedding propagation in KGs, we incorporate valuable relations using different triplet assumptions so that REP can maintain or mostly improve prediction quality.
    \item The proposed REP has significant scalability and takes less time to achieve comparable performance on large KGs, promising practical applications.
\end{itemize}

\begin{table*}[tbp]
    \centering
    \small
    \resizebox{\linewidth}{!}{
    \begin{tabular}{l c c c c} 
    \toprule
    \textbf{Method} & \textbf{Score Function} $f_r(h, t)$ & \textbf{Head Context Function} $g_h(h, r)$ & \textbf{Tail Context Function} $g_t(t, r)$ & \textbf{Parameters} \\
    \midrule
    TransE & $-\|\mathbf{h} + \mathbf{r} - \mathbf{t}\|$ & $\mathbf{h} + \mathbf{r}$ & $\mathbf{t} - \mathbf{r}$ & $\mathbf{h, r, t} \in \mathbb{R}^n$\\
    DistMult & $<\mathbf{r}, \mathbf{h}, \mathbf{t}>$ & $\mathbf{h} \odot \mathbf{r}$  & $\mathbf{t} \odot \mathbf{r}$ & $\mathbf{h, r, t} \in \mathbb{R}^n$ \\
    RotatE & $-\|\mathbf{h} \odot \mathbf{r} - \mathbf{t}\|$ & $\mathbf{h} \odot \mathbf{r}$ & $\mathbf{t} \odot \overline{\mathbf{r}}$ & $\mathbf{h, r, t} \in \mathbb{C}^n$ \\
    OTE & $\sum_{i=1}^{L} \|\text{diag}{(\text{exp}(\mathbf{s}_{r, (i)}))}\phi(\mathbf{M}_{r, (i)})\mathbf{h}_{(i)} - \mathbf{t}\|$ & $\sum_{i=1}^L \text{diag}{((\text{exp}(\mathbf{s}_{r, (i)}))}\phi(\mathbf{M}_{r, (i)})\mathbf{h}_{(i)}$ & $\sum_{i=1}^L \text{diag}{((\text{exp}(-\mathbf{s}_{r, (i)}))}\phi(\mathbf{M}_{r, (i)})^\top\mathbf{t}_{(i)}$ & $\mathbf{h, t} \in \mathbb{R}^n$ \\
    \bottomrule
    \end{tabular}
    }
    \caption{Score functions $f_r(h, t)$ and context functions $g_h(h, r)$ and $g_t(t, r)$ for different triplet-based KG embedding methods, where $< \cdot >$ denotes the generalized dot product, $\odot$ denotes the Hadamard product, $\overline{\cdot}$ denotes conjugate for complex vectors, $\phi$ denotes the Gram Schmidt process, $\mathbf{M}_r$ denotes orthogonal matrix for $r$, and $L$ denotes the number of parameter groups.}
    \label{tab:message_function}
\end{table*}

\section{Related Work} \label{sec:related_work}

\subsection{Knowledge Representation Learning}

Knowledge representation learning has been used to embed entities and relations in KGs into latent space and then infer missing facts based on existing ones.
The literature falls into two major categories:
(1) triplet-based methods that model triplet plausibility by score functions;
(2) context-based methods that model graph context beyond triplets.

Triplet-based methods often make assumptions about entities and relations in triplets.
Some regard relations as operations between connected entities in the latent space, such as addition \cite{TransE}, rotation \cite{RotatE}, and transformations in higher dimensional space \cite{QuatE,OTE}.
Others follow semantic matching and design score functions based on similarities \cite{DistMult,ComplEx}.
In general, these methods are of elegant simplicity and good interpretability.

Context-based methods usually inherit the message passing framework from GNNs \cite{MPNN}.
As heterogeneous relations always play an essential role in KGs, modeling such information is a promising direction.
GC-OTE \cite{OTE} expects the central embeddings close to the aggregation of neighbors with relation-specific orthogonal transformation.
R-GCN \cite{RGCN} extends GCN by introducing relation-specific weight matrices.
CompGCN \cite{CompGCN} follows the idea of modeling relations as composition operations.
However, it makes the same assumption about entity-relation pairs in both head and tail graph contexts, which requires external parameters to denote the direction of relations.

\subsection{Simplified Graph Neural Networks}

Recent progress in simplifying GNNs for homogeneous graphs has led to a surge of improvements in scalability.
SGC \cite{SGC} empirically observes that local averaging is more critical and removes non-linearity.
APPNP \cite{APPNP} utilizes a propagation schema derived from personalized PageRank and achieves linear computational complexity in the number of edges.
Reminiscent of our method, C\&S \cite{C&S} uses graph structure as a post-processing mechanism but propagates labels to correct predictions.

\begin{figure}[thp]
    \centering
    \includegraphics[width=0.98\columnwidth]{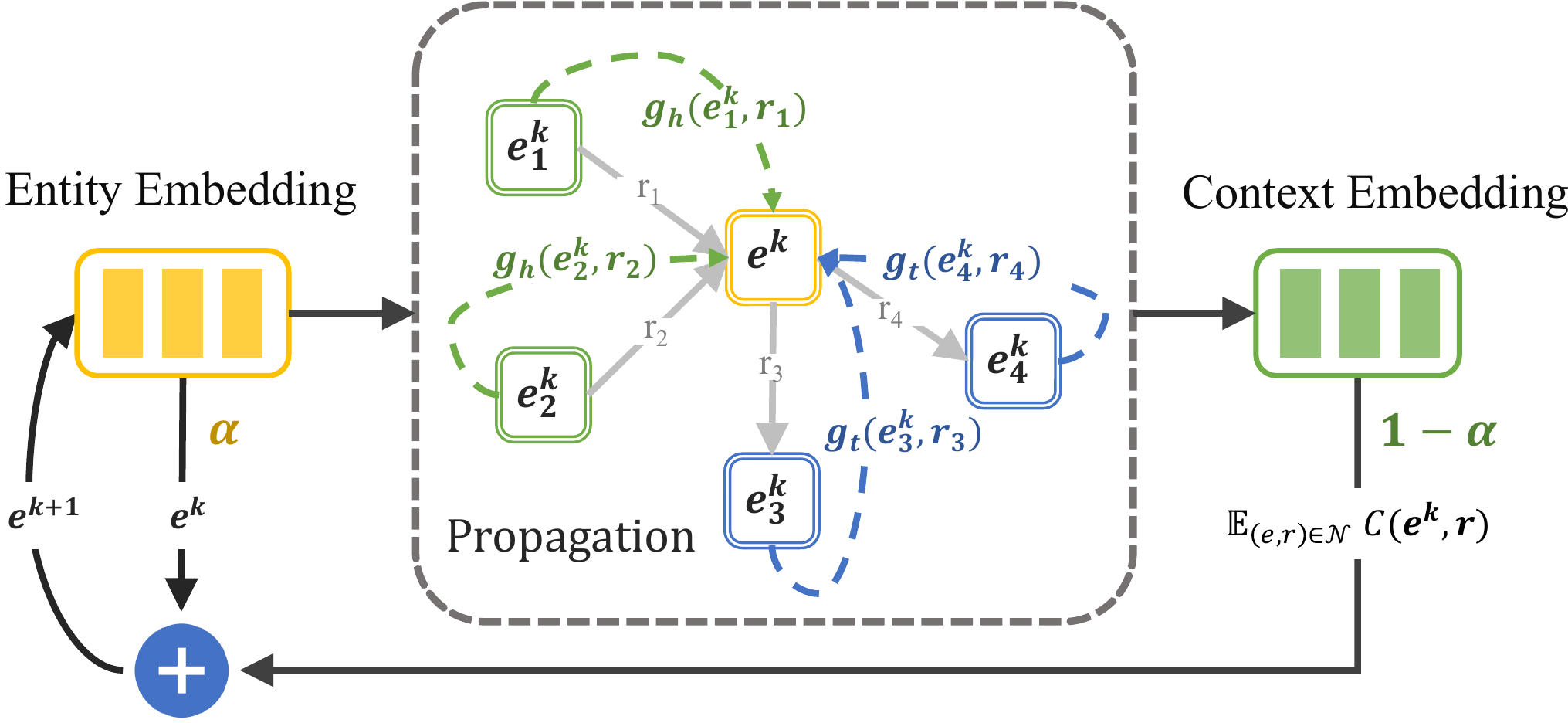}
    \caption{Overview of REP. It consists of two phases: (1) relation-based embedding propagation in the dashed box and (2) entity adaptation by weighted aggregation.} 
    \label{fig:framework}
\end{figure}

\section{Relation-based Embedding Propagation}

Our approach starts with simple triplet-based methods, which learn embeddings for entities and relations through minimizing objectives under triplet assumptions.
After, REP takes the pre-trained embeddings as input and improves them by incorporating neighbor information in graph context.
As illustrated in Figure \ref{fig:framework}, REP consists of
(1) relational graph context functions efficiently and effectively aggregate relation-specific neighbor information;
(2) the entity adaptation computes new embeddings based on pre-trained and context embeddings.

\paragraph{Notations.}
We consider knowledge graph as a collection of triplets $\mathcal{T} = \{(h, r, t)\ |\ h, t \in {E}, r \in {R}\} $ with ${E}$ as the entity set and ${R}$ as the relation set.
Let ${A}^H$ be the adjacent head list of the knowledge graph and ${A}^T$ be the adjacent tail list.
Specifically, the head adjacent list ${A}^H_i$ of the entity $i$ is composed of entity-relation pairs ($e, r$) with $(e_i, r, e) \in \mathcal{T}$. In contrast, the tail adjacent list ${A}^T_i$ contains pairs ($e, r$) with $(e, r, e_i) \in \mathcal{T}$.
Furthermore, corresponding bold letters denote embeddings.

\subsection{Relation-based Context Function} \label{sec:model}

A key motivation behind our relational context functions is that heterogeneous relations in KGs play a crucial role in understanding entities' meanings.
Triplet-based methods usually take relations as operations between entities such as additions, multiplications, rotations, and orthogonal transformations.
Thus, ignoring relations during propagation can bring information distortion.
We design relational context functions according to their assumptions to keep consistent with pre-trained embeddings.
As shown in Figure \ref{fig:framework}, there are two kinds of context.

\textit{Head Graph Context} includes incoming head-relation pairs for a central entity $e$, where $e$ acts as a tail entity.
As triplet-based methods often apply relation operations to head entities, we can directly use their assumptions for the head-relation pairs.
We choose four typical triplet-based methods, which regard relations as additions, multiplications, rotations, and orthogonal transformations, respectively, and define corresponding head context functions $g_h(h, r)$ in Table \ref{tab:message_function}.

\textit{Tail Graph Context} consists of all outgoing tail-relation pairs connected to the central entity.
KGs are directed graphs, where a valid $(h, r, t)$ does not mean $(t, r, h)$ is also a real fact.
Therefore, we can not use the same assumptions as of the head graph context.
Propagating information of tails and relations to heads utilize de facto inverse edges $(t, r^{-1}, h)$ in KGs.
These inverse edges are usually modeled through augmenting relations with inverse ones \cite{ConvE} or using external parameters to represent relations' directions \cite{RGCN}, which requires external computing resources.
To make our method concise and informative, we make opposite assumptions about the tail-relation pairs.
Specifically, embeddings of tails and relations are combined by the inverse operations of those for the head graph context, namely subtraction, multiplication, inverse rotation, and inverse orthogonal transformation.
In this way, the tail context functions $g_t(t, r)$ are defined in Table \ref{tab:message_function}.

Here we introduce how the context functions are derived in detail.
For the methods that minimize the distance between $\mathbf{t}$ and the combination of $\mathbf{h}$ and $\mathbf{r}$, we set their score functions to the desired 0.
Then we get $\|\mathbf{h} + \mathbf{r} - \mathbf{t}\| = 0$ for addition-based TransE, $\|\mathbf{h} \odot \mathbf{r} - \mathbf{t}\| = 0$ for rotation-based RotatE, and $\sum_{i=1}^{L} \|\text{diag}{(\text{exp}(\mathbf{s}_{r, (i)}))}\phi(\mathbf{M}_{r, (i)})\mathbf{h}_{(i)} - \mathbf{t}\| = 0$ for orthogonal transformation-based OTE.
After, we can deduce $\mathbf{h}$ and $\mathbf{t}$ from these equations.
In particular, the relation matrices in OTE are orthogonal so that we can get their inverse by simple transpose.
For DistMult that maximizes generalized dot product, we expect two elements as close as possible: either $\mathbf{t} = \mathbf{h} \odot \mathbf{r}$ or $\mathbf{h} = \mathbf{t} \odot \mathbf{r}$.

\textit{Context Aggregation} aims to combine neighbor information.
As this work aims to improve the scalability of graph context-based methods, we use averaging for aggregation, which helps to keep embeddings on the same scale.
Then the head context embedding $\Tilde{\mathbf{h}}_i^{(k)}$ of $e_i$ is defined as
\begin{equation}
    \Tilde{\mathbf{h}}_i^{(k)} = \frac{1}{|A_i^H|} C^H(\mathbf{e}_i^{(k)}) = \frac{1}{|A_i^H|} \sum_{e, r \in A^H_i}{g_h(\mathbf{e}^{(k)}, r)},
\end{equation}
and the tail context embedding $\Tilde{\mathbf{t}}_i^{(k)}$ is
\begin{equation}
    \Tilde{\mathbf{t}}_i^{(k)} = \frac{1}{|A_i^T|} C^T(\mathbf{e}_i^{(k)}) = \frac{1}{|A_i^T|} \sum_{e, r \in A^T_i}{g_t(\mathbf{e}^{(k)}, r)}.
\end{equation}

\subsection{Entity Adaptation}

The propagation schema derived from personalized PageRank \cite{APPNP} can be formulated as Eq. (\ref{eq:appnp}).
\begin{equation} \label{eq:appnp}
    \mathbf{E}^{(k+1)} = (1 - \alpha)\mathcal{A}\mathbf{E}^{(k)} + \alpha \mathbf{E}^{(0)},
\end{equation}
where $\mathbf{E}^{(0)}$ is the input feature matrix, and $\mathcal{A}$ is the symmetrically normalized adjacency matrix with self-loops.
Here we denote embeddings after $k$ updates as $\mathbf{E}^{(k)}$.

However, storing both $\mathbf{E}^{(k)}$ and $\mathbf{E}^{(0)}$ is very demanding on memory for large KGs.
We do not store the input feature matrix $\mathbf{E}^{(0)}$, i.e., pre-trained embeddings in our method, but use them for embedding initialization.
Besides, we separate self-loops from $\mathcal{A}$ and use different weights for entity and context embeddings to retain inherent information in pre-trained embeddings.
Especially, Eq. (\ref{eq:appnp}) can not utilize relations in KGs, so we design relational context functions in previous Section \ref{sec:model}.

In summary, we use an update scalar $\alpha \in [0, 1)$ to balance the trade-off between the pre-learned triplet information and the graph context information.
Then the final embedding of entity $e_i$ is computed according to Eq. (\ref{eq:REP}).
\scalebox{0.96}{\parbox{1.04\linewidth}{
\begin{equation}\label{eq:REP}
    \mathbf{e}_i^{(k+1)} = \alpha \mathbf{e}_i^{(k)} + \frac{1 - \alpha}{|A^H_i| + |A^T_i|} (C^H(\mathbf{e}_i^{(k)}) + C^T(\mathbf{e}_i^{(k)})),
\end{equation}
}}
where $\mathbf{e}^{(0)}$ is the pre-trained entity embeddings.
In particular, instead of updating parameters with gradient descent algorithms, we directly use the computation results for link prediction.
As the number of relations is relatively small, we fix the pre-trained embeddings of heterogeneous relations.

\subsection{{Theoretical Analysis}} \label{sec:provement}

In this section, we provide theoretical analysis from the parameter update perspective.
Generally, triplet-based methods start with their score function $f_r(h, t)$ and learn embeddings through minimizing the margin-based ranking criterion:
\begin{align} \label{eq:transe}
        \mathcal{L} & = -\mathbb{E}_{(h,r,t) \in \mathcal{T}}[\gamma + f_r(\mathbf{h}, \mathbf{t}) - f_r(\mathbf{h}', \mathbf{t}')]_{+},
\end{align}
where $\gamma$ denotes the parameter margin, $[x]_{+}$ takes the positive part of $x$. A negative sample $(h', r, t')$ is constructed by uniformly replacing $h$ or $t$ in $(h, r, t)$ with other entities in $E$.
Parameters of embeddings are updated by gradient descent methods. 
The typical stochastic gradient descent algorithm can be formulated as Eq. (\ref{eq:sgd}).
\begin{equation} \label{eq:sgd}
    \mathbf{e}_i' = \mathbf{e}_i - \beta \frac{\partial \mathcal{L}}{\partial e_i},
\end{equation}
where $\beta$ denotes the learning rate.
The motivation behind the margin-based ranking criterion is to make valid triplets have relatively higher scores by enlarging the distance to negative samples.
Therefore, when embeddings are fully optimized by this criterion, the distances between valid triplets and negative samples reach a local optimum.

\begin{proof}

Here we prove that our REP further improves the objective $\mathcal{L}$ by maximizing scores of valid triplets.
Without loss of generality, we take TransE as an example. It interprets relations as additions in latent space, i.e., $\mathbf{t}$ is expected to be the nearest neighbor of $\mathbf{h} + \mathbf{r}$.
Accordingly, the score function is defined as $f_r(h, t)=-\|\mathbf{h} + \mathbf{r} - \mathbf{t}\|$.
Then the objective of maximizing scores of valid triplets is formulated as Eq. (\ref{eq:loss_pos}).

\begin{equation} \label{eq:loss_pos}
    \mathcal{J} = -\mathbb{E}_{(h_i, r_i, t_i) \in \mathcal{T}} -\| \mathbf{h}_i + \mathbf{r}_i - \mathbf{t}_i \|.
\end{equation}
Assuming that parameters are updated by stochastic gradient descent method, we first calculate the partial derivative for $h_i$ as shown in Eq. (\ref{eq:transe_grad}).
\begin{align} \label{eq:transe_grad}
    \frac{\partial \mathcal{J}}{\partial h_i} &= 2 \mathbb{E}_{(h_j, r_j, t_j) \in \mathcal{T} \land (h_i=h_j)} (\mathbf{h}_j + \mathbf{r}_j - \mathbf{t}_j ) \nonumber \\
    &= 2 \mathbf{h}_i + 2 \mathbb{E}_{(h_j, r_j, t_j) \in \mathcal{T} \land (h_i=h_j)} (\mathbf{r}_j - \mathbf{t}_j).
\end{align}

In this case, Eq. (\ref{eq:sgd}) can be reformulated as Eq. (\ref{eq:final_h}) by replacing the partial derivative with Eq. (\ref{eq:transe_grad}).
\begin{equation} \label{eq:final_h}
    \mathbf{h}_i' = (1 - 2 \beta)\mathbf{h}_i + 2 \beta \mathbb{E}_{(h_j, r_j, t_j) \in \mathcal{T} \land (h_i=h_j)} (\mathbf{t}_j - \mathbf{r}_j ).
\end{equation}
Let $\alpha = 1 - 2 \beta$, this update function is equivalent to Eq. (\ref{eq:REP}) when entities act as heads, where the head context function is $\mathbf{t} - \mathbf{r}$.
For the tail entity $t_i$, we have
\begin{align} \label{eq:final_t}
    \mathbf{t}_j'
    &= (1 - 2 \beta)\mathbf{t}_j + 2 \beta \mathbb{E}_{(h_j, r_j, t_j) \in \mathcal{T} \land (t_i=t_j)} (\mathbf{h}_j + \mathbf{r}_j),
\end{align}
where the addition operation is consistent with the tail context function in Table \ref{tab:message_function}.
Above all, REP acts the same as using stochastic gradient descent to maximize valid triplets' scores, which further optimizes the triplet-based objective $\mathcal{L}$ of pre-trained embeddings.
\end{proof}

\begin{table}[tp]
    \centering
    \small
    \resizebox{\linewidth}{!}{
    \begin{tabular}{l r r r r }
    \toprule
    {\textbf{Dataset}} & {\# \textbf{Entities}} & {\# \textbf{Relations}} & \# \textbf{Triplets} \\
    \midrule
    WN18RR  & 40,943 & 11 & 86,835 \\
    FB15k-237 & 14,541 & 237 & 272,115 \\
    OGBL-WikiKG2  & 2,500,604 & 535 & 16,109,182 \\
    WikiKG90M-LSC & 87,143,637 & 1,315 & 504,220,369 \\
    \bottomrule
    \end{tabular}
    }
    \caption{Statistics of knowledge graphs on link prediction.}
    \label{tab:dataset}
\end{table}

\section{Experiments}

\subsection{Datasets}

We use four datasets at diverse scales, as reported in Table \ref{tab:dataset}.

\paragraph{Large scale.} 
WikiKG90M-LSC \cite{OGB-LSC} of KDD Cup 2021 \footnote{https://ogb.stanford.edu/kddcup2021/} is extracted from Wikidata. 
Aside from triplets, it provides text features learned by RoBERTa \cite{Roberta}, which are combined by concatenation when training triplet-based methods. 
It also provides 1000 randomly sampled entity candidates for each tail in the validation/test set.

\paragraph{Medium scale.} 
OGBL-WikiKG2 \cite{OGB} is a link prediction dataset from Open Graph Benchmark (OGB) \footnote{https://ogb.stanford.edu/}.
It is constructed from the relational statements in Wikidata after filtering out rare entities.
Moreover, 1000 randomly sampled candidates for tail entities are provided likewise.

\paragraph{Small scale.} 
FB15k-237 and WN18RR \cite{ConvE} are the most widely used datasets for link prediction.
FB15k-237 mostly describes people-related facts, while WN18RR consists of hyponym and hypernym relations.

\subsection{Evaluation Metrics} 
Link prediction aims to predict {missing} facts based on existing triplets, namely to predict valid heads for $(t, r)$ or valid tails for $(h, r)$.
Specifically, we first corrupt test triplets and construct candidate triplets using entity candidates.
As small datasets do not provide candidates, we use all entities $e$ that $(h, r, e) \notin \mathcal{T}$ and $(e, r, t) \notin \mathcal{T}$ as candidates for a test triplet ($h, r, t$).
Then we compute the plausibility of the test triplet and its candidate triplets and sort them by descending order.
Evaluation metrics are based on the rank of test triplets, including Mean Reciprocal Rank (MRR) and Hits@K which denote the proportion of ranks less than K.

\begin{table}[tp]
    \small
    \centering
    \begin{tabular}{c c c c}
    \toprule
    \textbf{Dataset} & \textbf{REP-OTE} & \textbf{GC-OTE} & \textbf{Speedup} \\
    \midrule
    WN18RR & 16s & 19s & 1.2 \\
    FB15k-237 & 96s & 1959s & 20.4\\
    OGBL-WikiKG2 & 1.7h & $>$ 100h & $>$ 50 \\
    \bottomrule
    \end{tabular}
    \caption{Time complexity comparison between REP-OTE and GC-OTE. The values denote the time cost for each epoch.}
    \label{tab:time}
\end{table}

\subsection{Implementation Details}
Experiments were conducted on Intel Xeon Gold 6271C CPUs and Tesla V100 SXM2 GPUs.
We use the public code to reproduce triplet-based methods, while the results of context-based methods are from original papers.
The code and details of REP are released in Graph4KG
\footnote{https://github.com/PaddlePaddle/PGL/tree/main/apps/Graph4KG}.

\subsection{Results and Discussion}

\begin{table*}[thp]
    \centering
    \small
    \begin{tabular}{l c c c c c | c c c c}
    \toprule
    \multirow{2}{*}{\textbf{Base Model}} & \multirow{2}{*}{\textbf{Strategy}} & \multicolumn{4}{c}{\textbf{OGBL-WikiKG2}} & \multicolumn{4}{c}{\textbf{WikiKG90M-LSC}} \\
    \cmidrule{3-10}
    & & MRR & Hits@1 & Hits@3 & Hits@10 & MRR & Hits@1 & Hits@3 & Hits@10 \\
    \midrule
    \multirow{2}{*}{TransE \cite{TransE}} & - &  0.426 & 0.397 & 0.429 & 0.468 & 0.881 & 0.845 & 0.908 & \textbf{0.947} \\
    & REP & \textbf{0.528} & \textbf{0.481} & \textbf{0.542} & \textbf{0.619} & \textbf{0.884} & \textbf{0.850} & \textbf{0.912} & \textbf{0.947}\\
    \multirow{2}{*}{DistMult \cite{DistMult}} & - & 0.369 & 0.322 & 0.374 & 0.453  & 0.863 & 0.834 & 0.885 & 0.918 \\
    & REP &  \textbf{0.376} & \textbf{0.329} & \textbf{0.380} & \textbf{0.461} & \textbf{0.868} & \textbf{0.838} & \textbf{0.891} & \textbf{0.925} \\
    \multirow{2}{*}{RotatE \cite{RotatE}} & - & 0.434 & 0.399 & 0.441 & 0.491 & 0.886 & 0.850 & 0.914 & 0.952 \\
    & REP & \textbf{0.491} & \textbf{0.438} & \textbf{0.507} & \textbf{0.593} & \textbf{0.889} & \textbf{0.854} & \textbf{0.916} & \textbf{0.953}\\
    \multirow{2}{*}{OTE \cite{OTE}} & - & 0.519 & 0.475 & 0.532 & 0.597 & 0.910 & 0.881 & 0.933 & 0.961\\
    & REP & \textbf{0.538} & \textbf{0.490} & \textbf{0.554} & \textbf{0.624} & \textbf{0.914} & \textbf{0.884} & \textbf{0.938} & \textbf{0.966}\\
    \bottomrule
    \end{tabular}
    \caption{Evaluation results on the medium-scale  (OGBL-WikiKG2) and large-scale (WikiKG90M-LSC) KGs on link prediction.}
    \label{tab:wikikg2}
\end{table*}

\begin{table*}[!th]
    \small
    \centering
    \begin{tabular}{l c c c c | c c c c }
    \toprule
    \multirow{2}{*}{\textbf{Model}} & \multicolumn{4}{c}{\textbf{FB15k-237}} & \multicolumn{4}{c}{\textbf{WN18RR}} \\ \cmidrule{2-9}
     & MRR & Hits@1 & Hits@3 & Hits@10  & MRR & Hits@1 & Hits@3 & Hits@10 \\
    \midrule
    TransE \cite{TransE}  & 0.329 & 0.230 & 0.368 & {0.528} & 0.226 & 0.016 & {0.403} & {0.531} \\
    DistMult \cite{DistMult} & 0.307 & 0.219 & 0.336 & 0.483 & 0.440 & 0.394 & 0.454 & 0.532 \\
    RotatE \cite{RotatE} & 0.336	 & 0.240 & {0.374} & {0.531} & {0.477} & {0.432} & {0.494} & {0.567}  \\
    OTE \cite{OTE} & 0.352 & 0.258 & 0.387 & 0.539 & 0.487 & 0.438 & 0.502 & {0.587} \\
    PairRE \cite{PairRE} & 0.351 & 0.256 & 0.387 & 0.544 & 0.454 & 0.411 & 0.469 & 0.548 \\
    \midrule
    R-GCN \cite{RGCN} & 0.248 & 0.151 & - & 0.417 & - & - & - & - \\
    SACN \cite{SACN} & 0.350 & 0.260 & 0.390 & 0.540 & 0.470 & 0.430 & 0.480 & 0.540 \\
    TransGCN \cite{TransGCN} & 0.356 & 0.252 & 0.388 & {0.555} & 0.485 & 0.438 & {0.510} & 0.578 \\
    CompGCN \cite{CompGCN} & 0.355 & 0.264 & 0.390 & 0.535 & 0.479 & {0.443} & 0.494 & 0.546 \\
    {GC-OTE} \cite{OTE} & {0.360} & {0.265} & \textbf{0.398} & {0.547} & \textbf{0.491} & {0.442} & \textbf{0.512} & {0.583} \\
    M$^2$\text{GNN} \cite{M2GNN} & \textbf{0.362} & \textbf{0.275} & \textbf{0.398} & \textbf{0.565} & 0.485 & \textbf{0.444} & 0.498 & 0.572 \\
    \midrule
    REP-OTE & {0.354} & {0.262} & {0.388} & {0.540} & {0.488} & {0.439} & {0.505} & \textbf{0.588}  \\
    \bottomrule
    \end{tabular}
    \caption{Prediction results of different models on FB15k-237 and WN18RR. The best results are bold.}
    \label{tab:baseline}
\end{table*}

\paragraph{Simplicity.}
For model parameters, REP requires no parameters except for embeddings of entities and relations, while context-based methods inherit from GNNs and require multiple layer-specific weight matrices besides embeddings.
For the computation complexity, we conduct speed experiments to compare REP-OTE and GC-OTE. GC-OTE also requires no external parameters.
In theory, REP-OTE and GC-OTE have the same time complexity $\mathcal{O}(|\mathcal{T}|\mathcal{D})$, where $|\mathcal{T}|$ is the number of triplets and $\mathcal{D}$ denotes the average degree.
However, as REP has no backward propagation, its constant is far smaller than GC-OTE.
The speedup numbers in Table \ref{tab:time} empirically prove this.
We count the time consumed to traverse all training data.
To make a fair comparison, the time of REP-OTE consists of the time REP takes and the time used by OTE in an epoch during pre-training.
We can note that REP brings more acceleration when the data scale is larger.

\paragraph{Effectiveness.}
Results of medium and large datasets are reported in Table \ref{tab:wikikg2}.
We have observed that when there are more than 80K entities and 1.6M triplets, context-based methods like CompGCN will be out of memory on a 32GB GPU to learn embeddings with dimension 200.
Thus, we only report their results on small-scale datasets.
As we can see, REP-augmented methods perform best on all metrics.
In particular,  REP-TransE significantly surpasses TransE and is competitive with OTE, the state-of-the-art triplet-based method with higher complexity.
It averagely achieves a 10\% relative improvement on OGBL-WikiKG2 compared with four triplet-based baselines.
This result may suggest that graph context is of great use for knowledge graphs at a medium scale.
The improvement of WikiKG90M-LSC is not as significant as that on OGBL-WikiKG2.
We reason that the text features used during pre-training provide external information to entities with few triplets, so graph context is possible to be superfluous.
Results on small datasets are reported in Table \ref{tab:baseline}.
In this case, REP-OTE outperforms all triplet-based methods and has comparable prediction quality to the state-of-the-art context-based methods.
Above all, REP can bring stable improvements to triplet-based methods and achieve competitive performance as context-based methods with less cost.

\begin{figure}[!tp]
    \centering
    \includegraphics[width=0.98\columnwidth]{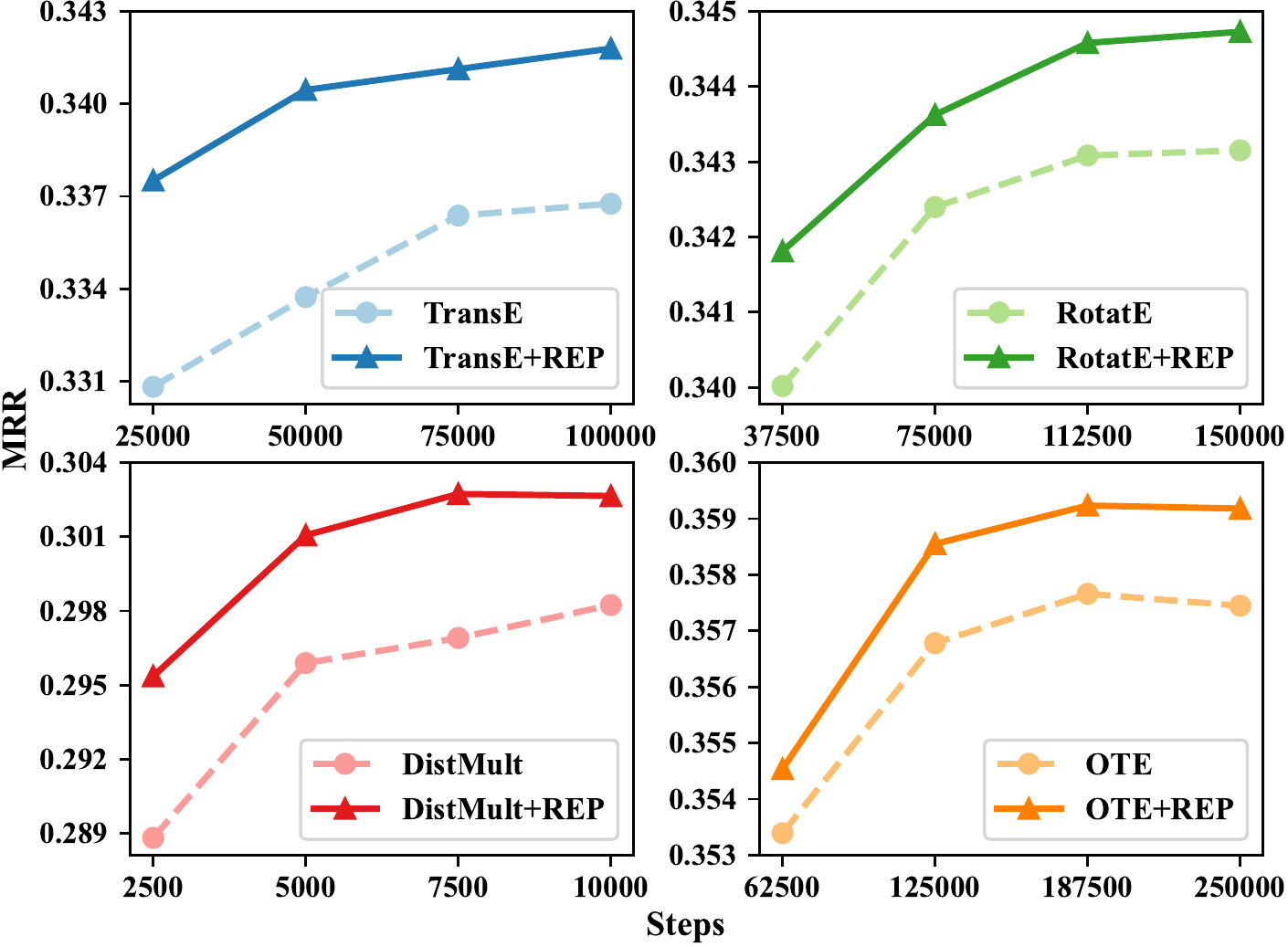}
    \caption{The MRR performance of REP on four triplet-based methods at different training stages.}
    \label{fig:unconverge-OTE}
\end{figure}

\paragraph{Practical Effect.}
We assume in previous experiments that triplet-based methods have fully trained pre-trained embeddings.
In this section, we explore how REP performs on not converged embeddings.
Specifically, we apply REP to the embeddings trained by triplet-based methods after 0.25N, 0.5N, 0.75N, and N steps separately if the embeddings converge at step N. 
The MRR results of these embeddings are reported in Figure \ref{fig:unconverge-OTE}.
Experiments on the four triplet-based methods show that REP brings a significant performance boost to not converged embeddings.
Particularly, triplet-based methods with REP can achieve comparable results as the converged ones with just half of the training steps.
In other words, REP allows obtaining high-quality embeddings within less time.

\begin{figure*}[tp]
    \centering
    \begin{subfigure}{0.245\textwidth}
        \centering
        \includegraphics[width=\linewidth]{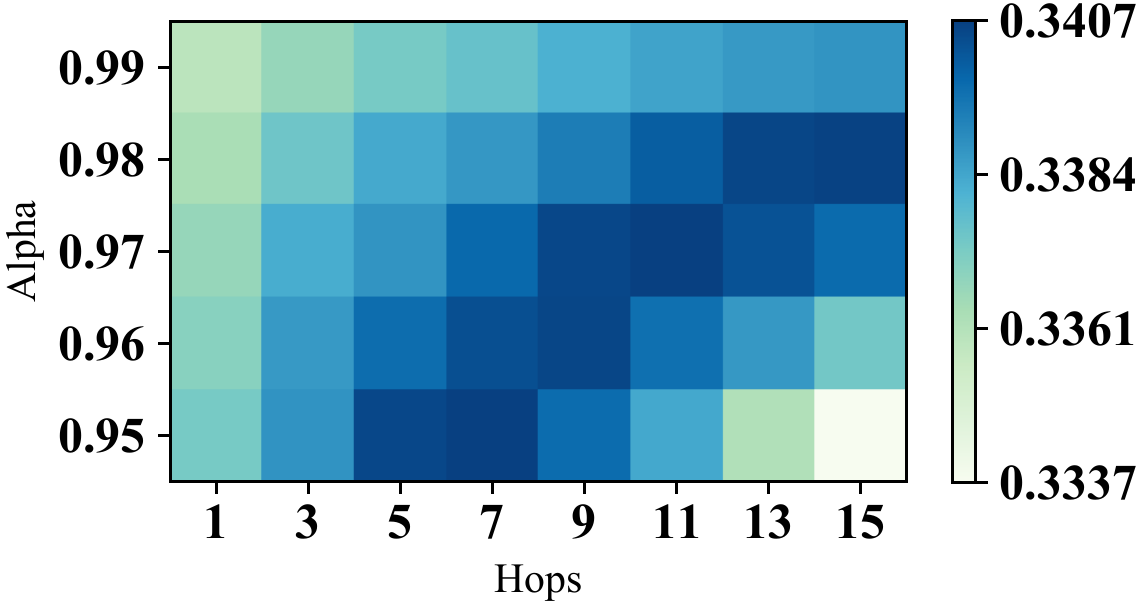}
        \caption{TransE}
        \label{fig:transe_alpha}
    \end{subfigure}
    \begin{subfigure}{0.245\textwidth}
        \centering
        \includegraphics[width=\linewidth]{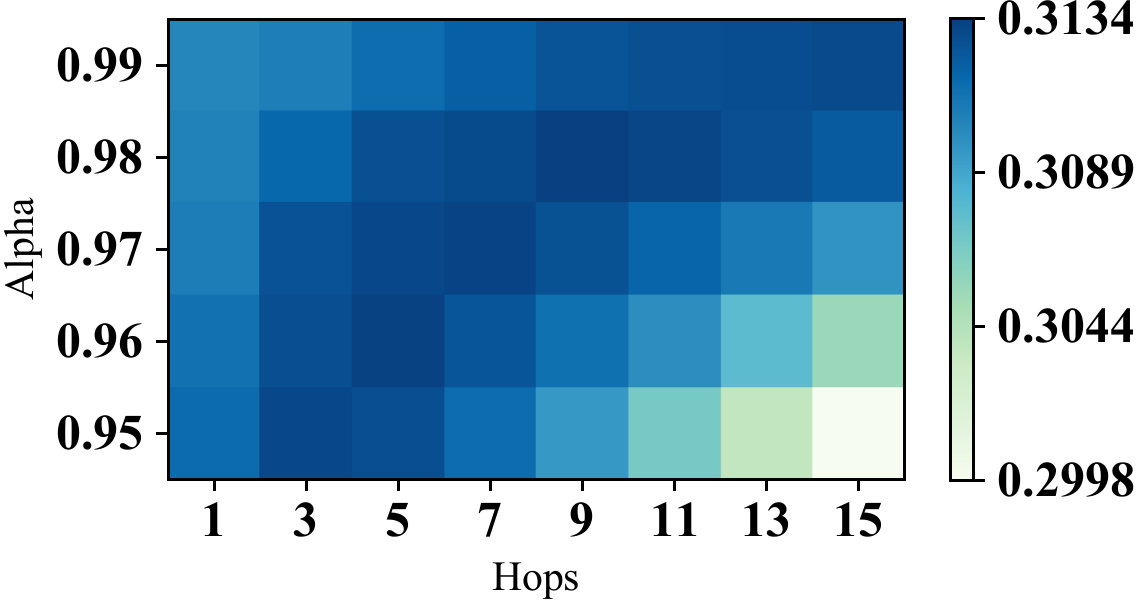}
        \caption{DistMult}
        \label{fig:distmult_alpha}
    \end{subfigure}
    \begin{subfigure}{0.245\textwidth}
        \centering
        \includegraphics[width=\linewidth]{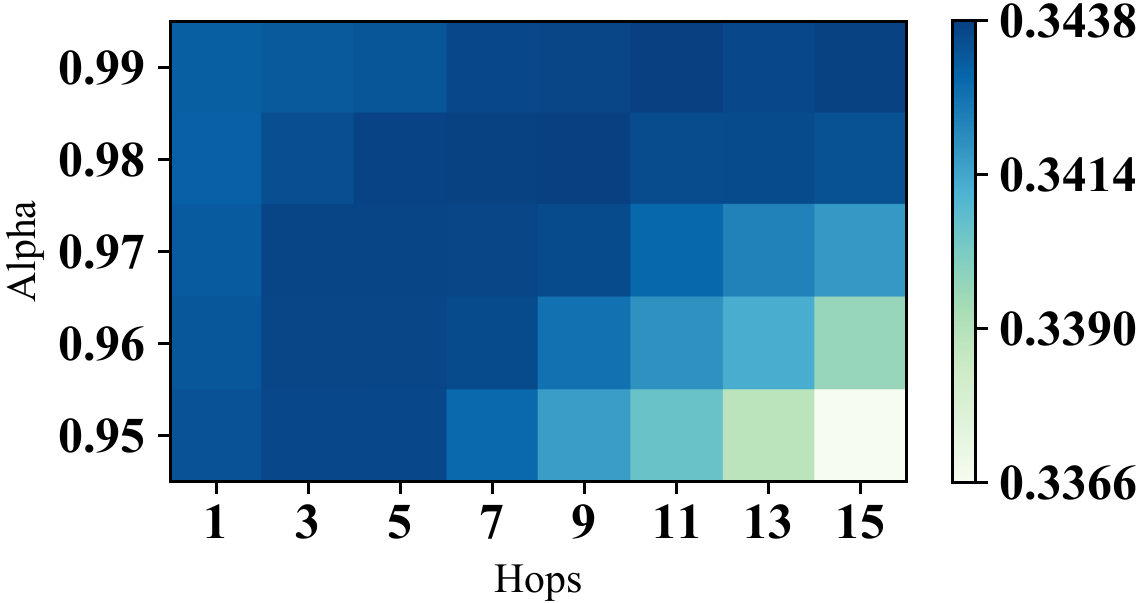}
        \caption{RotatE}
        \label{fig:rotate_alpha}
    \end{subfigure}
    \begin{subfigure}{0.245\textwidth}
        \centering
        \includegraphics[width=\linewidth]{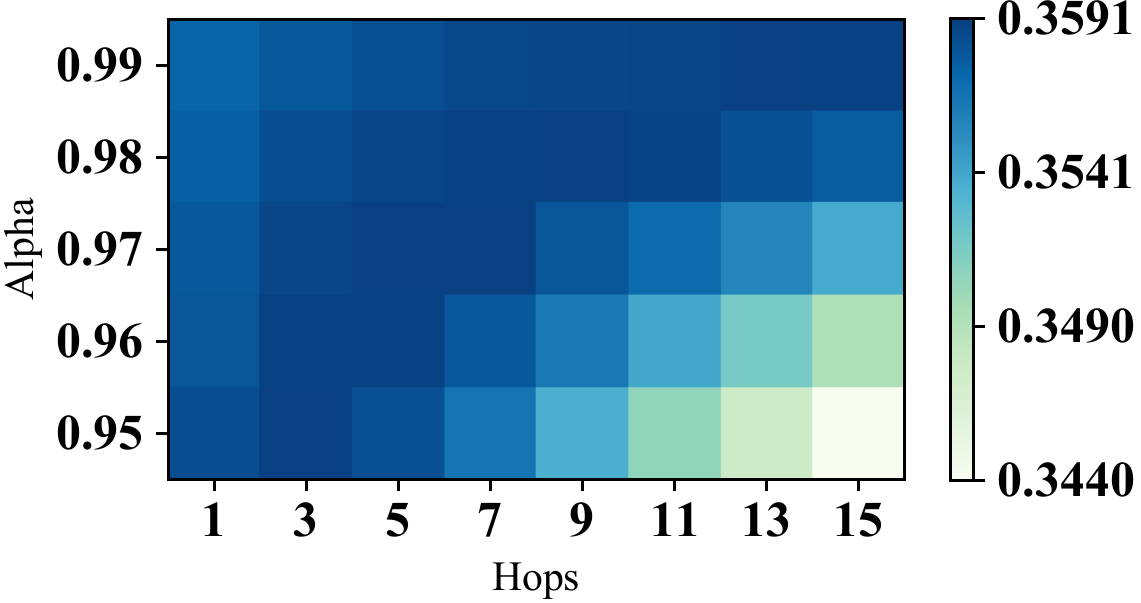}
        \caption{OTE}
        \label{fig:ote_alpha}
    \end{subfigure}
    \caption{MRR results of triplet-based methods plus REP on FB15k-237 with different hyper-parameters.}
    \label{fig:ablation_alpha}
\end{figure*}

\begin{figure*}[!ht]
    \centering
    \begin{subfigure}{0.25\textwidth}
        \centering
        \includegraphics[width=\linewidth]{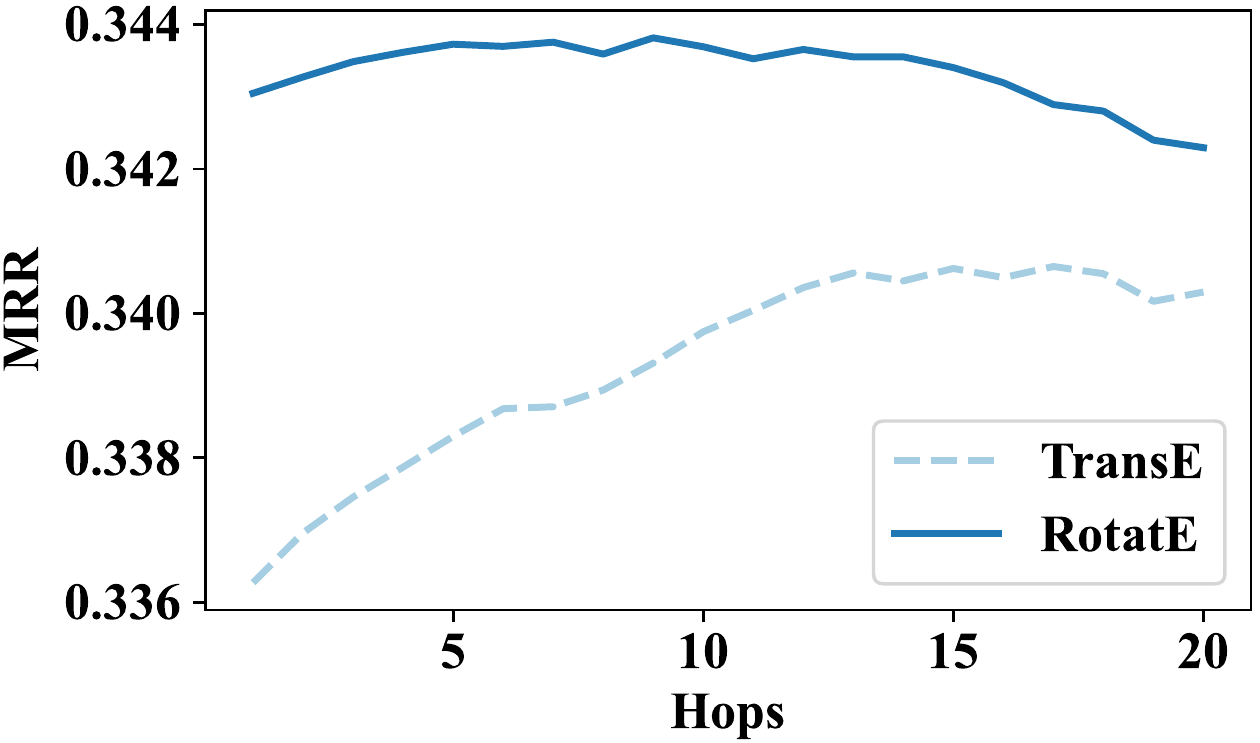}
        \caption{FB15k-237}
        \label{fig:fb_hop}
    \end{subfigure}
    \quad\quad\quad
    \begin{subfigure}{0.25\textwidth}
        \centering
        \includegraphics[width=\linewidth]{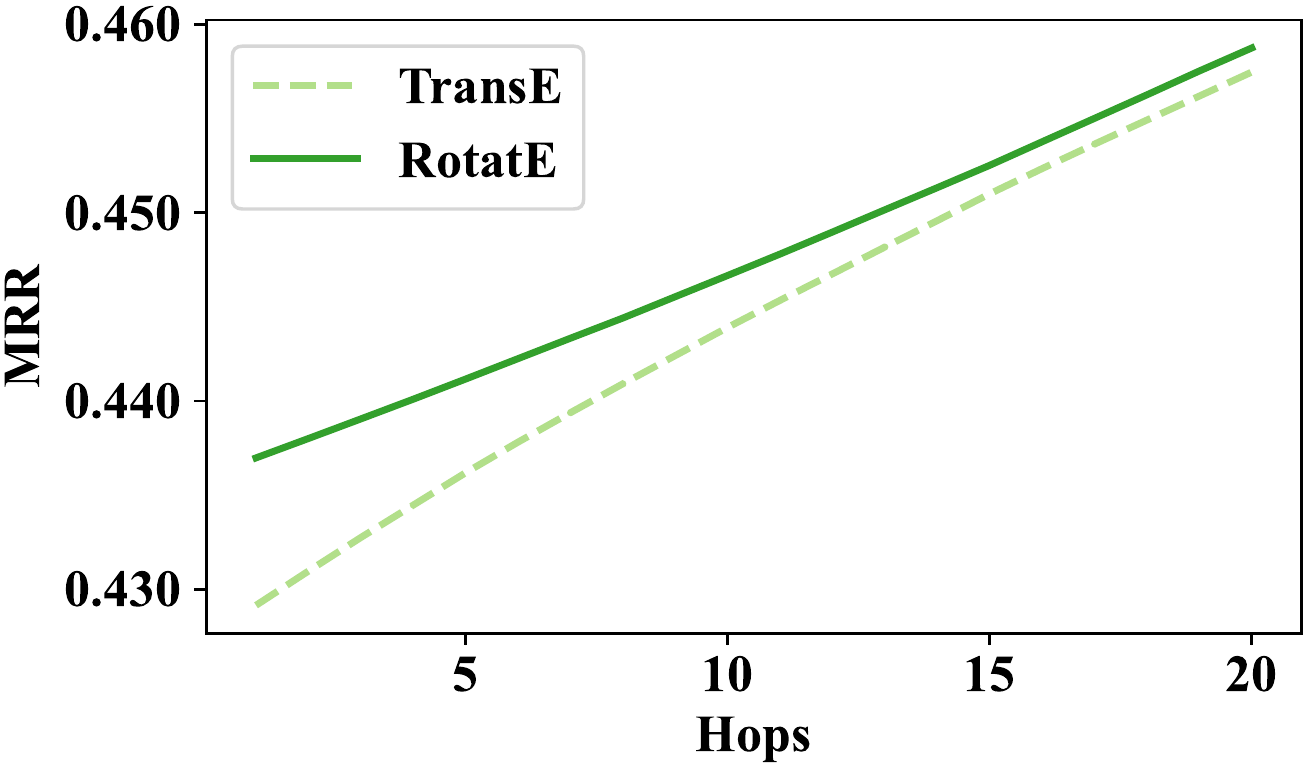}
        \caption{OGBL-WikiKG2}
        \label{fig:wiki_hop}
    \end{subfigure}
    \quad\quad\quad
    \begin{subfigure}{0.25\textwidth}
        \centering
        \includegraphics[width=\linewidth]{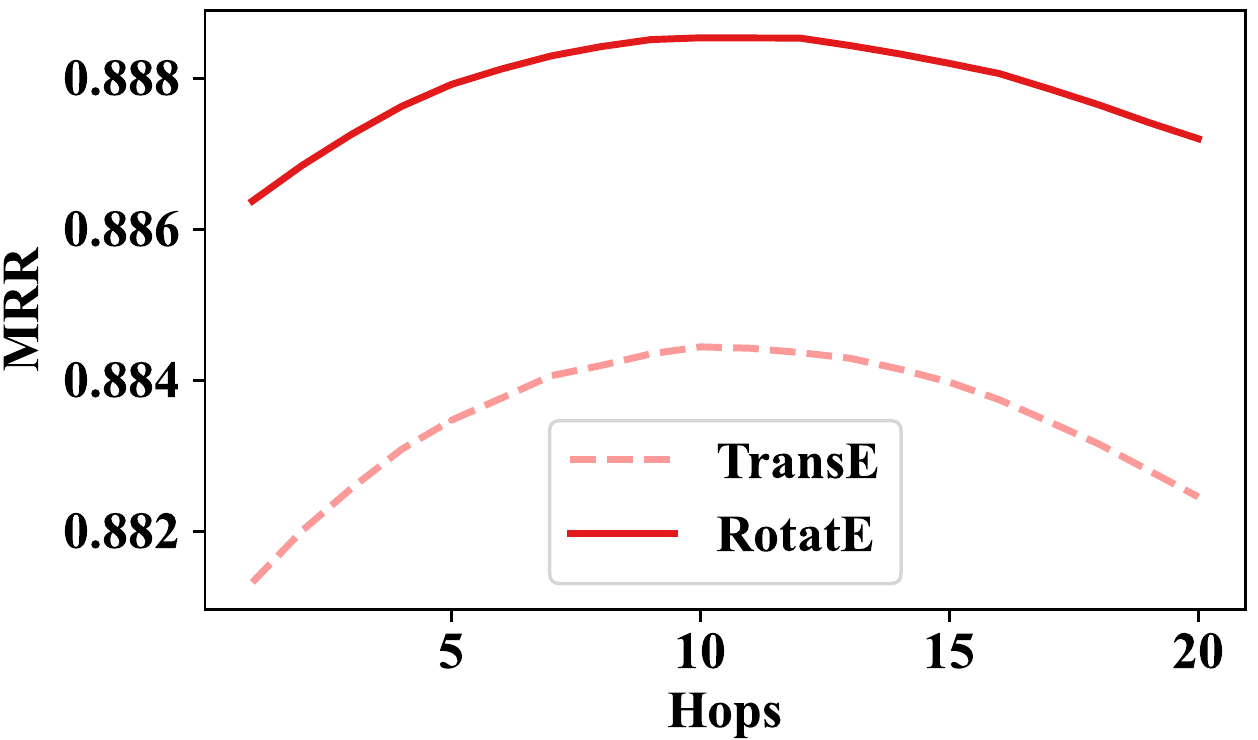}
        \caption{WikiKG90M-LSC}
        \label{fig:wiki90m_hop}
    \end{subfigure}
    \caption{Results of triplet-based methods plus REP at different hops on datasets on three different scales.}
    \label{fig:ablation_hop}
\end{figure*}

\begin{figure}[!thp]
    \centering
    \includegraphics[width=0.98\columnwidth]{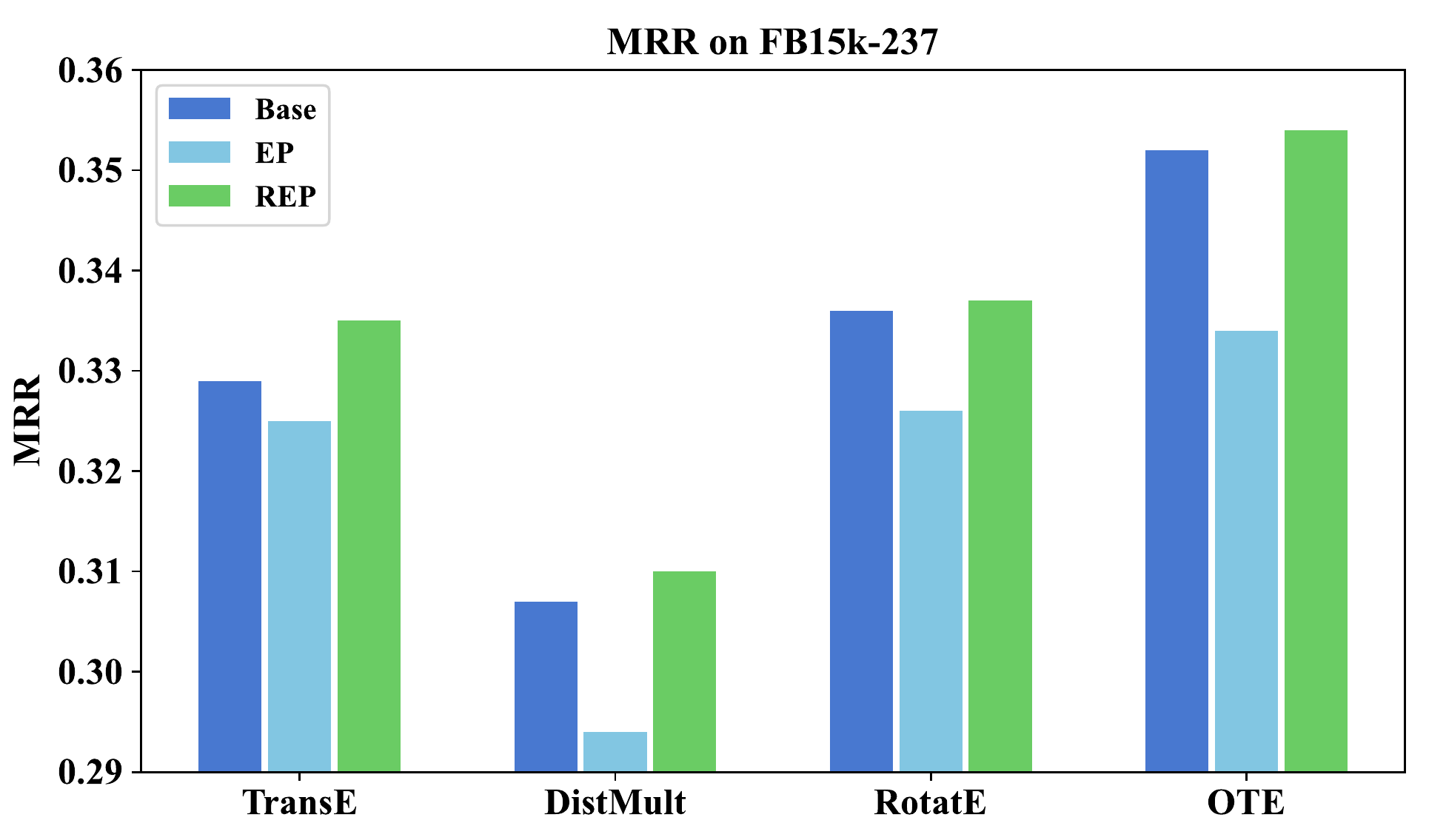}
    \caption{Results of triplet-based methods and the corresponding plus embedding propagation (EP) methods and plus REP methods on the FB15k-237 dataset.}
    \label{fig:ablation_relation}
\end{figure}

\paragraph{How to select $\alpha$?} The update scalar $\alpha$ determines how much neighbor information contributes to the central entities. 
To investigate the impact of $\alpha$, we set $\alpha$ as values ranging from 0.95 to 0.99 with step 0.01.
Then we evaluate embeddings after once to 15 times entity adaptations separately and draw heat maps using their MRRs in Figure \ref{fig:ablation_alpha}.
As illustrated in Section \ref{sec:provement}, $\alpha = 1 - 2\beta$, where $\beta$ denotes learning rate.
Figure \ref{fig:ablation_alpha} also demonstrates this point.
When $\alpha$ increases, REP needs more iterations to get the best MRR.

\paragraph{Are multi-hop neighbors always helpful?} 
For REP, the central entity receives neighbor information k-hops away after $k$ times entity adaption.
We investigate how multi-hop neighbors influence embedding quality by changing the number of hops from 1 to 20.
We set $\alpha=0.98$ to control variables.
Results of TransE and RotatE on datasets at different scales are plotted in Figure~\ref{fig:ablation_hop}.
As we can see, MRRs of OGBL-WikiKG2 grow almost linearly as the number of hops increases, maybe because of its triplet sparsity.
Moreover, the curves of both methods on FB125k-237 and WikiKG90M-LSC are nearly concave.
It shows that enlarging the receptive field contributes to embedding quality, but neighbors too far away would hurt performance when triplets are dense, or there is external information.

\subsection{Influence of Relations}

We have theoretically analyzed the necessity of relations in Section \ref{sec:provement}. 
In this section, we provide some empirical evidence.
\textit{Embedding Propagation} (EP) methods like APPNP have shown strong performance on homogeneous graphs with single relation.
However, when applied to KGs, they can not utilize types of relations and regard relations as identity matrices.
In order to prove the significance of relations, we conduct ablation studies on the FB15k-237 dataset.
Figure \ref{fig:ablation_relation} plots the evaluation results of four triplet-based methods. {Base} denotes MRRs of fully trained triplet-based methods.
{EP} propagates entity embeddings to neighbor entities directly and ignores relations during this process.
We also apply EP to the pre-trained embeddings as our {REP} method.
As shown in Figure \ref{fig:ablation_relation}, embedding propagation without relations (EP) causes significant performance deterioration.
It proves that the relation-based context functions designed in Section \ref{sec:model} play a critical role during embedding propagation.

\section{Conclusion}

We proposed the novel method REP to utilize graph context in KGs during the post-training process.
The key idea is incorporating relational graph structure information into pre-trained triplet-based embeddings.
For simplicity, we use local averaging to conduct non-parametric embedding propagation without backward propagation.
For effectiveness, we design graph context functions for head-relation neighbors and tail-relation neighbors, respectively.
As a result, REP can reduce information distortion during embedding propagation in KGs.
Furthermore, such design brings REP excellent scalability, which has practical significance in utilizing graph context for large KGs in the real world.
Experimental results also show that by enriching pre-trained triplet-based embeddings with graph context, REP improves or maintains prediction quality with less time cost.

\section*{Acknowledgments}

This work has been supported by National Key Research and Development Program of China (2018AAA0101900).

\bibliographystyle{named}
\bibliography{ijcai22}

\end{document}